\documentclass{article}

\usepackage{microtype}
\usepackage{graphicx}
\usepackage{subcaption}
\usepackage{booktabs}
\usepackage[hidelinks]{hyperref}
\usepackage{amsmath}
\usepackage{amssymb}
\usepackage{mathtools}
\usepackage{amsthm}
\usepackage[capitalize,noabbrev]{cleveref}
\usepackage[disable,textsize=tiny]{todonotes}
\usepackage{enumitem}

\usepackage{tikz}
\usetikzlibrary{shapes.geometric, arrows.meta, positioning, fit, backgrounds, calc, decorations.pathreplacing}

\PassOptionsToPackage{numbers, sort&compress}{natbib}
\usepackage[preprint]{neurips_2026}

\usepackage{amsmath,amsfonts,bm}

\def\eqref#1{equation~\ref{#1}}

\def\1{\bm{1}}

\def\vf{{\bm{f}}}

\def\vh{{\bm{h}}}

\def\vx{{\bm{x}}}
\def\vy{{\bm{y}}}

\DeclareMathAlphabet{\mathsfit}{\encodingdefault}{\sfdefault}{m}{sl}
\SetMathAlphabet{\mathsfit}{bold}{\encodingdefault}{\sfdefault}{bx}{n}

\theoremstyle{plain}

\theoremstyle{definition}

\theoremstyle{remark}

\graphicspath{{images/}}

\title{Recommending Bacterial Cultivation Conditions \\
from 16S rRNA}
\author{%
  Juntang Wang\thanks{Equal contribution.} \\
  Harvard University \\
  USA \\
  \And
  Hao Wu\footnotemark[1] \\
  Duke Kunshan University \\
  China \\
  \And
  Yihan Wang\footnotemark[1] \\
  Duke University \\
  USA \\
  \AND
  Runkun Guo \\
  Boston University \\
  USA \\
  \And
  Shixin Xu \\
  Duke Kunshan University \\
  China \\
  \And
  Dongmian Zou\thanks{Corresponding author: \texttt{dongmian.zou@dukekunshan.edu.cn}.} \\
  Duke Kunshan University \\
  China \\
}

\begin{document}

\maketitle

\begin{abstract}
Recommending bacterial cultivation conditions from sequence data is a challenging structured problem characterized by missing standardized benchmarks and limited use of sequence structure.
To address this, we introduce DSNI, a benchmark of 65,023 bacterial strains with phylogeny-aware splits, pairing 16S rRNA sequences with four cultivation conditions (media, temperature, pH, oxygen). Exploiting the mosaic structure of 16S rRNA, where hypervariable regions support discrimination and conserved regions support primer design, we propose a Variability-Gated Encoder that learns position-dependent importance weights conditioned on biological features (region type, position, and local complexity). The gating module adds minimal overhead ($1{,}280$ parameters, ${<}1\%$) while providing an explicit inductive bias valuable in low-data regimes. Experiments demonstrate that the learnable gate improves flat Transformers by 4.7\% in joint accuracy and complements RNABERT, yielding 44.2\% exact-match joint accuracy. The model also recovers 90.2\% of ground-truth cultivation protocols within its top-25 recommendations, narrowing the exhaustive search space by approximately 60$\times$.

\end{abstract}

\section{Introduction}
\label{sec:intro}

Leveraging domain structure to improve generalization in low-data regimes is a central challenge in representation learning. 16S ribosomal RNA \citep{woesePhylogeneticStructureProkaryotic1977} offers a compelling testbed for this challenge due to its distinctive mosaic structure: nine hyper-variable regions (V1--V9) carrying discriminative signals, interspersed with highly conserved segments \citep{chakravortyDetailedAnalysis16S2007}. While hyper-variable regions enable species-level discrimination and conserved regions support universal PCR primer design, prior work has largely treated 16S as a taxonomic marker rather than a predictor of bacterial phenotypes. 
We ask whether models can use 16S sequences to recommend cultivation conditions, including growth media, temperature, pH, and oxygen requirements, that microbiologists must empirically determine for novel organisms. This structured recommendation task presents three technical challenges: (1) scarce labeled data relative to sequence length, (2) severe class imbalance (89\% mesophiles, 91\% neutrophiles), and (3) output multiplicity (organisms may tolerate multiple conditions). These challenges motivate incorporating inductive bias: can we encode domain knowledge about sequence structure into neural architectures to improve sample efficiency?

Cultivation is essential for characterizing microbial diversity, yet more than 99\% of microbial diversity remains uncultured, a phenomenon termed the ``great plate count anomaly'' \citep{staleyMEASUREMENTSITUACTIVITIES1985, rappeUnculturedMicrobialMajority2003}.
The primary barrier is the laborious trial-and-error process of identifying appropriate growth conditions for each novel organism\citep{lagierCulturingHumanMicrobiota2018, lewisInnovationsCulturingUncultured2021}. If cultivation conditions could be predicted from sequence data alone, this bottleneck could be substantially relieved, turning exhaustive screening into a targeted shortlist and opening sequence-guided cultivation as a concrete route toward studying the uncultured majority.

We hypothesize that 16S sequences correlate with cultivation requirements due to the phylogenetic conservation of metabolic traits \citep{martinyMicrobiomesLightTraits2015, gobernaPredictingMicrobialTraits2016}. While 16S does not directly encode metabolic enzymes, its sequence reflects phylogenetic history that shapes phenotypic similarity. We frame the task as \emph{structured recommendation}, using sequence-derived patterns to suggest plausible cultivation conditions. Validating this hypothesis requires: (i) a large-scale dataset pairing 16S with standardized cultivation metadata, and (ii) methods for learning cultivation-relevant representations. Currently, no such benchmark or dedicated method exists.

To address this gap, we introduce \textbf{DSNI}, a benchmark curated from DSMZ \citep{reimerBacDive20222022,schoberBacDive20252025} DSNI contains 65{,}023 strains, roughly $3.6\times$ the size of the closest comparable resource KOMODO \citep{oberhardtHarnessingLandscapeMicrobial2015}. 
To our knowledge, it is the first 16S benchmark that pairs sequences with all four primary cultivation conditions under a single standardized labeling scheme, providing a unified resource for sequence-based cultivation recommendation at scale.
To support rigorous evaluation, DSNI uses phylogeny-aware splits that cluster sequences by taxonomic similarity. This protocol prevents information leakage from near-neighbor organisms and more realistically evaluates generalization to novel taxa.

Beyond DSNI, we propose a \textbf{Variability-Gated Encoder}: a lightweight gating mechanism ($1{,}280$ parameters, $<$1\% overhead) that learns position-dependent importance weights conditioned on region type (V1--V9 vs.\ conserved), normalized position, and local sequence complexity. This allows the model to allocate representational capacity based on domain-informed features rather than hard-coding rules or relying solely on attention to discover these patterns. We benchmark our approach against flat encoders and pretrained models (RNABERT), using identical decoders and training budgets to isolate the gating effect and test whether domain-informed bias complements pretraining.

Our contributions are three-fold: (1) a 65{,}023-strain 16S benchmark with standardized four-condition labels and phylogeny-aware splits; (2) a lightweight Variability-Gated Encoder with biologically grounded inductive bias; (3) systematic experiments showing that gating improves joint accuracy by +4.7\%, complements RNABERT to reach 44.2\%, and achieves 90.2\% top-25 protocol recall, reducing the candidate protocol space by ${\sim}60\times$.

\section{Related Work}
\label{sec:related}

\textbf{16S Databases.}
Large-scale 16S rRNA databases, including SILVA \citep{quastSILVARibosomalRNA2012}, Greengenes \citep{desantisGreengenesChimeraChecked16S2006}, and RDP \citep{coleRibosomalDatabaseProject2014}, primarily support taxonomic classification \citep{janda16SRRNAGene2007}. Although 16S has been evaluated for species- and strain-level resolution \citep{johnsonEvaluation16SRRNA2019a}, these resources are not designed for cultivation-oriented prediction. PICRUSt \citep{langillePredictiveFunctionalProfiling2013} infers functional gene content from 16S via ancestral state reconstruction, but targets genomic potential rather than cultivation requirements.

\textbf{Cultivation Condition Prediction.}
The most related prior work is KOMODO \citep{oberhardtHarnessingLandscapeMicrobial2015}, which predicts growth media by collaborative filtering on phylogenetically similar strains and reports 83\% accuracy on organism-media pairings under random train/test splits. However, random splits can place closely related strains in both train and test sets, leading to phylogenetic leakage. Under our genus-level splits, the corresponding GROWREC method achieves 46.3\% media accuracy (\Cref{tab:main_results}). 
Other work predicts optimal growth temperature from genome features \citep{sauerPredictingOptimalGrowth2019a} or growth rates from codon usage \citep{weissmanEstimatingMaximalMicrobial2021}, but they require complete genomes or rely on close phylogenetic references. To our knowledge, no standardized benchmark or task-specific model exists for predicting bacterial cultivation conditions directly from 16S rRNA sequences.

\textbf{Sequence Representation Learning.}
Deep learning for nucleic acid sequences began with convolutional models for binding prediction \citep{alipanahiPredictingSequenceSpecificities2015} and has since evolved into pretrained foundation models, including DNABERT \citep{jiDNABERTPretrainedBidirectional2021}, Nucleotide Transformer \citep{dalla-torreNucleotideTransformerBuilding2025}, HyenaDNA \citep{nguyenHyenaDNALongRangeGenomic2023}, and RNA-specific models such as RNABERT \citep{akiyamaInformativeRNABase2022}, RNA-FM \citep{chenInterpretableRNAFoundation2022}, and ERNIE-RNA \citep{yinERNIERNARNALanguage2025}. These models provide strong baselines, but they lack task-specific inductive biases. Unlike Highway Networks \citep{srivastavaHighwayNetworks2015} and Gated CNNs \citep{dauphinLanguageModelingGated2017a}, where gates are learned from internal representations, our gating mechanism conditions on external biological features (region type and local complexity) and provides explicit inductive bias.

\section{The DSNI Benchmark}
\label{sec:dataset}

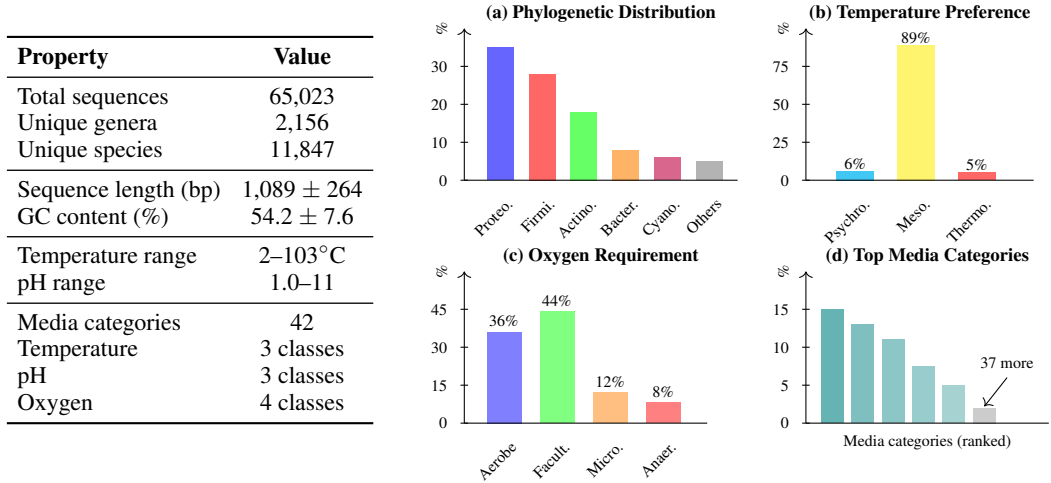
\begin{figure}[!t]
    \centering
    \begin{minipage}[t]{0.36\textwidth}
        \vspace{1.5em}
        \centering
        \small
        \setlength{\tabcolsep}{4pt}
        \begin{tabular}{lc}
            \toprule
            \textbf{Property} & \textbf{Value} \\
            \midrule
            Total sequences & 65,023 \\
            Unique genera & 2,156 \\
            Unique species & 11,847 \\
            \midrule
            Sequence length (bp) & 1,089 $\pm$ 264 \\
            GC content (\%) & 54.2 $\pm$ 7.6 \\
            \midrule
            Temperature range & 2--103$^\circ$C \\
            pH range & 1.0--11 \\
            \midrule
            Media categories & 42 \\
            Temperature & 3 classes \\
            pH & 3 classes \\
            Oxygen & 4 classes \\
            \bottomrule
        \end{tabular}
    \end{minipage}\hfill
    \begin{minipage}[t]{0.60\textwidth}
        \vspace{0pt}
        \centering
        \resizebox{\linewidth}{!}{%

\begin{tikzpicture}[
    bar/.style={draw=none, minimum width=0.35cm},
    xlabel/.style={font=\tiny, rotate=45, anchor=north east},
    ylabel/.style={font=\tiny},
    title/.style={font=\scriptsize\bfseries},
    legendbox/.style={font=\tiny},
]

\begin{scope}[shift={(0, 3.2)}]
    \node[title] at (1.8, 2.2) {(a) Phylogenetic Distribution};

    \draw[->] (0, 0) -- (0, 2) node[ylabel, rotate=90, anchor=south, yshift=1mm] {\%};
    \foreach \y/\lab in {0/0, 0.5/10, 1/20, 1.5/30} {
        \draw (-0.05, \y) -- (0.05, \y);
        \node[ylabel, anchor=east] at (-0.1, \y) {\lab};
    }

    \fill[blue!60] (0.3, 0) rectangle (0.65, 1.75);   
    \fill[red!60] (0.85, 0) rectangle (1.2, 1.4);     
    \fill[green!60] (1.4, 0) rectangle (1.75, 0.9);   
    \fill[orange!60] (1.95, 0) rectangle (2.3, 0.4);  
    \fill[purple!60] (2.5, 0) rectangle (2.85, 0.3);  
    \fill[gray!60] (3.05, 0) rectangle (3.4, 0.25);   

    \node[xlabel] at (0.55, -0.05) {Proteo.};
    \node[xlabel] at (1.1, -0.05) {Firmi.};
    \node[xlabel] at (1.65, -0.05) {Actino.};
    \node[xlabel] at (2.2, -0.05) {Bacter.};
    \node[xlabel] at (2.75, -0.05) {Cyano.};
    \node[xlabel] at (3.3, -0.05) {Others};

    \draw (0, 0) -- (3.6, 0);
\end{scope}

\begin{scope}[shift={(4.5, 3.2)}]
    \node[title] at (1.5, 2.2) {(b) Temperature Preference};

    \draw[->] (0, 0) -- (0, 2) node[ylabel, rotate=90, anchor=south, yshift=1mm] {\%};
    \foreach \y/\lab in {0/0, 0.5/25, 1/50, 1.5/75} {
        \draw (-0.05, \y) -- (0.05, \y);
        \node[ylabel, anchor=east] at (-0.1, \y) {\lab};
    }

    \fill[cyan!60] (0.4, 0) rectangle (0.9, 0.12);      
    \fill[yellow!70] (1.2, 0) rectangle (1.7, 1.78);    
    \fill[red!60] (2.0, 0) rectangle (2.5, 0.1);        

    \node[xlabel] at (0.75, -0.05) {Psychro.};
    \node[xlabel] at (1.55, -0.05) {Meso.};
    \node[xlabel] at (2.35, -0.05) {Thermo.};

    \draw (0, 0) -- (3.0, 0);

    \node[font=\tiny] at (0.65, 0.22) {6\%};
    \node[font=\tiny] at (1.45, 1.88) {89\%};
    \node[font=\tiny] at (2.25, 0.2) {5\%};
\end{scope}

\begin{scope}[shift={(0, 0)}]
    \node[title] at (1.8, 2.2) {(c) Oxygen Requirement};

    \draw[->] (0, 0) -- (0, 2) node[ylabel, rotate=90, anchor=south, yshift=1mm] {\%};
    \foreach \y/\lab in {0/0, 0.5/15, 1/30, 1.5/45} {
        \draw (-0.05, \y) -- (0.05, \y);
        \node[ylabel, anchor=east] at (-0.1, \y) {\lab};
    }

    \fill[blue!50] (0.3, 0) rectangle (0.75, 1.2);      
    \fill[green!50] (1.0, 0) rectangle (1.45, 1.47);    
    \fill[orange!50] (1.7, 0) rectangle (2.15, 0.4);    
    \fill[red!50] (2.4, 0) rectangle (2.85, 0.27);      

    \node[xlabel] at (0.6, -0.05) {Aerobe};
    \node[xlabel] at (1.3, -0.05) {Facult.};
    \node[xlabel] at (2.0, -0.05) {Micro.};
    \node[xlabel] at (2.7, -0.05) {Anaer.};

    \draw (0, 0) -- (3.2, 0);

    \node[font=\tiny] at (0.52, 1.35) {36\%};
    \node[font=\tiny] at (1.22, 1.62) {44\%};
    \node[font=\tiny] at (1.92, 0.55) {12\%};
    \node[font=\tiny] at (2.62, 0.42) {8\%};
\end{scope}

\begin{scope}[shift={(4.5, 0)}]
    \node[title] at (1.6, 2.2) {(d) Top Media Categories};

    \draw[->] (0, 0) -- (0, 2) node[ylabel, rotate=90, anchor=south, yshift=1mm] {\%};
    \foreach \y/\lab in {0/0, 0.5/5, 1/10, 1.5/15} {
        \draw (-0.05, \y) -- (0.05, \y);
        \node[ylabel, anchor=east] at (-0.1, \y) {\lab};
    }

    \fill[teal!60] (0.2, 0) rectangle (0.5, 1.5);    
    \fill[teal!50] (0.6, 0) rectangle (0.9, 1.3);    
    \fill[teal!45] (1.0, 0) rectangle (1.3, 1.1);    
    \fill[teal!40] (1.4, 0) rectangle (1.7, 0.75);   
    \fill[teal!35] (1.8, 0) rectangle (2.1, 0.5);    
    \fill[gray!40] (2.2, 0) rectangle (2.5, 0.2);    

    \draw (0, 0) -- (3.2, 0);

    \node[font=\tiny] at (1.6, -0.25) {Media categories (ranked)};

    \draw[<-, thin] (2.35, 0.22) -- (2.65, 0.6) node[font=\tiny, anchor=south] {37 more};
\end{scope}

\end{tikzpicture}%
        }
    \end{minipage}

    \caption{DSNI benchmark statistics and label distributions. Left: summary statistics for 65,023 strains. Right: phylogenetic, temperature, oxygen, and media distributions. DSNI is dominated by Proteobacteria and Firmicutes, shows strong mesophilic bias, and has long-tailed media categories.}
    \label{fig:dataset_overview}
\end{figure}

We introduce DSNI (\textbf{DS}MZ and \textbf{N}IH \textbf{I}ntegrated), a curated benchmark for recommending bacterial cultivation conditions from 16S rRNA sequences. DSNI integrates records from DSMZ and NIH/NCBI resources and provides standardized labels for growth media, temperature, pH, and oxygen requirements. We use 16S as input because it is broadly recoverable across taxa and environments, making the benchmark applicable to organisms for which complete genome data is unavailable (Appendix \ref{app:why_16s} provides a quantitative comparison with genome features).

\subsection{Data Sources and Processing}
\label{sec:data_sources}

\textbf{Data Sources.}
DSNI integrates records from DSMZ \citep{reimerBacDive20222022} and NIH/NCBI resources \citep{sayersDatabaseResourcesNational2025}. DSMZ provides curated bacterial strain records with detailed cultivation protocols, while NIH/NCBI resources provide 16S sequences in GenBank and cultivation data from associated repositories, including BEI Resources. We cross-reference strain identifiers to link sequences with cultivation conditions.

\textbf{Database Construction.}
The combined records comprise six relational tables (STRAINS, MEDIA, SOLUTIONS, INGREDIENTS, STEPS, GAS) capturing taxonomic information, cultivation protocols, and precise media compositions. We flatten this schema into supervised sequence-condition pairs. Full schema details and entity-relationship diagrams are provided in Appendix \ref{app:dataset_schema}.

\textbf{Quality Control and Deduplication.}
Sequences are quality-filtered ($500$--$1{,}500$\,bp, $<$1\% ambiguous bases, SILVA-aligned~\citep{quastSILVARibosomalRNA2012}) and dereplicated at the 97\% OTU threshold~\citep{yarzaUnitingClassificationCultured2014}, retaining the entry with the most complete metadata per cluster; full pipeline in Appendix \ref{app:qc_dedup}.

\subsection{Label Standardization and Dataset Statistics}
\label{sec:dataset_stats}

\textbf{Label Standardization.}
Growth media names varied across sources, such as ``Tryptic Soy Agar'', ``TSA'', and ``Trypticase Soy Agar''. We standardized media through rule-based synonym matching and expert curation, with canonical categories defined by base composition, such as peptone-based, blood-based, and minimal-salts media. This preserved biologically meaningful distinctions, including aerobic versus anaerobic variants, while collapsing trivial naming differences. Rare media with fewer than 10 occurrences were grouped into an ``Other'' category. For continuous cultivation conditions, raw temperatures span 2$^\circ$C to 103$^\circ$C and pH values span 1.0 to 11. We discretize temperature into psychrophilic ($T < 20^\circ$C), mesophilic ($20^\circ\text{C} \leq T \leq 45^\circ$C), and thermophilic ($T > 45^\circ$C); pH into acidophilic (pH $< 6$), neutrophilic ($6 \leq \text{pH} \leq 8$), and alkaliphilic (pH $> 8$). Oxygen requirements are standardized into obligate aerobe, facultative anaerobe, microaerophile, and obligate anaerobe.

\textbf{Dataset Statistics}
\Cref{fig:dataset_overview} summarizes key statistics and the label distributions. The dataset exhibits class imbalance characteristic of cultured bacterial collections. Raw temperature values cluster heavily in the 20--40$^\circ$C range, reflecting the predominance of mesophilic organisms; after binning, mesophiles constitute 89\% of samples, with psychrophiles (6\%) and thermophiles (5\%, including rare hyperthermophilic bacteria from extreme environments) as minority classes. Similarly, raw pH values concentrate between 6 and 8, yielding 91\% neutrophiles after discretization. Most organisms are facultative anaerobes (44\%) or obligate aerobes (36\%), with microaerophiles (12\%) and obligate anaerobes (8\%) less represented. Media categories follow a long-tail distribution: the top 5 media (including Tryptic Soy, Brain Heart Infusion, and Nutrient Agar variants) account for 51\% of samples, while 19 media categories have fewer than 100 examples each. This imbalance motivates class-weighted losses and macro-averaged evaluation metrics.

\subsection{Phylogeny-Aware Splits}
\label{sec:splits}

To evaluate generalization to novel taxa, we construct splits at the genus level. Using NCBI Taxonomy, we ensure all sequences from a given genus exclusively in one split (train, validation, or test), preventing models from memorizing genus-specific signatures. We first sort genera by size, then greedily assign each genus to minimize deviation from the target proportions (70\%/15\%/15\%) while preserving stratification over the four label distributions. The final splits contain 1,509/324/323 genera and 45,516/9,754/9,753 sequences for train/validation/test. This procedure helps maintain representation of minority classes in validation and test sets.

\subsection{Task Definition and Evaluation Metrics}
\label{sec:task}

We frame cultivation condition recommendation as a multi-task classification problem. Given a 16S rRNA sequence $\vx = (x_1, \ldots, x_L)$ where $x_i \in \{A, C, G, U\}$ and $L$ denotes the sequence length, the goal is to predict $\vy = (y^{\text{media}}, y^{\text{temp}}, y^{\text{pH}}, y^{\text{O}_2})$: media category (42 classes), temperature preference (psychrophile/mesophile/thermophile), pH preference (acidophile/neutrophile/alkaliphile), and oxygen requirement (aerobe/facultative/microaerophile/anaerobe).

While we formalize this as a multi-task classification problem for learning, the underlying task is better understood as \emph{structured recommendation} rather than causal prediction. Models should learn to suggest plausible conditions from sequence-derived patterns in the training data. 
We note that cultivation conditions are interdependent. For example, thermophilic organisms often require specialized media formulations, motivating treating the four outputs jointly rather than as independent predictions. Multiple valid cultivation protocols may exist for a given organism. When source records indicate multiple valid media for a single strain, we retain only the primary recommended medium (typically the first listed in DSMZ protocols or the most frequently cited in NCBI records) as the ground-truth label. This single-label formulation simplifies training, while top-$k$ metrics account for near-miss predictions on alternative valid media.

\textbf{Per-Task Accuracy.}
For each task (media, temperature, pH, oxygen), we report top-1 accuracy on the test set. This provides an interpretable measure of how often the model's most confident prediction matches the ground truth. For media prediction, where multiple valid cultivation media may support growth of the same organism, we additionally report top-3 accuracy to credit near-miss predictions.

\textbf{Macro-F1 Score.}
Given the severe class imbalance in DSNI, we report macro-averaged F1 scores that weight each class equally regardless of frequency. This ensures that rare but biologically important classes contribute to the overall metric, following standard practice for imbalanced classification.

\textbf{Joint Accuracy.}
We report the fraction of test examples where \emph{all four} condition predictions are correct. This stringent metric measures the model's ability to produce coherent cultivation protocols, reflecting the reality that a single incorrect condition can lead to culture failure. Joint accuracy penalizes models that perform well on individual tasks but fail to capture inter-task correlations.

\section{Variability-Gated Encoder}
\label{sec:model}

We propose a Variability-Gated Encoder that incorporates biologically motivated inductive bias. The key insight is that hypervariable regions (V1--V9) provide species-level discrimination, while conserved regions reflect evolutionary constraints. Rather than hard-coding region weights or relying solely on attention to infer positional importance from limited data, we introduce a learnable gating layer conditioned on externally computed biological features. This provides an explicit inductive bias: the model is told \emph{which} features may matter, but learns \emph{how} they matter through end-to-end training.

To handle the ${\sim}1{,}500$ bp span of 16S rRNA sequences, we employ a Transformer backbone for long-range sequence modeling. The proposed gate adds position-wise importance conditioned on biological features that attention cannot directly encode, which is valuable in low-data regimes. The module is efficient, adding only $1{,}280$ parameters ($<$1\% overhead), and is consistent with prior evidence that gating can complement attention \citep{dalla-torreNucleotideTransformerBuilding2025}.


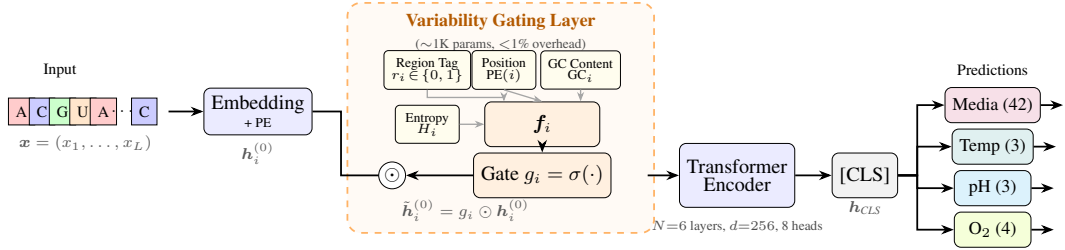
\begin{figure*}[!t]
    \centering
    \resizebox{\textwidth}{!}{%
    \begin{tikzpicture}[
        box/.style={draw, rounded corners=3pt, minimum height=0.7cm, align=center, font=\small},
        mainbox/.style={box, fill=blue!8, minimum width=1.6cm},
        gatebox/.style={box, fill=orange!12, minimum width=1.5cm},
        headbox/.style={box, fill=green!10, minimum width=1.2cm, minimum height=0.55cm, font=\footnotesize},
        featurebox/.style={draw, rounded corners=2pt, fill=yellow!8, minimum height=0.42cm, minimum width=1.0cm, font=\tiny, align=center},
        seqbox/.style={draw, minimum height=0.42cm, minimum width=0.32cm, font=\scriptsize},
        arrow/.style={-{Stealth[length=2mm]}, thick},
        grayarrow/.style={-{Stealth[length=1.5mm]}, gray!70, semithick},
        label/.style={font=\scriptsize, align=center},
        eqlabel/.style={font=\scriptsize\itshape, text=gray!70!black},
        annot/.style={font=\tiny, text=gray!60!black},
    ]
    
    \node[label] at (-5.6, 0.65) {Input};
    \node[seqbox, fill=red!20] (x1) at (-6.2, 0) {A};
    \node[seqbox, fill=blue!20] (x2) at (-5.88, 0) {C};
    \node[seqbox, fill=green!20] (x3) at (-5.56, 0) {G};
    \node[seqbox, fill=orange!20] (x4) at (-5.24, 0) {U};
    \node[seqbox, fill=red!20] (x5) at (-4.92, 0) {A};
    \node[font=\scriptsize] at (-4.6, 0) {$\cdots$};
    \node[seqbox, fill=blue!20] (xL) at (-4.28, 0) {C};
    \node[eqlabel] at (-5.2, -0.5) {$\vx = (x_1, \ldots, x_L)$};
    
    \node[mainbox] (embed) at (-2.5, 0) {Embedding\\[-2pt]{\tiny + PE}};
    \draw[arrow] (-3.9, 0) -- (embed);
    \node[eqlabel] at (-2.5, -0.65) {$\vh_i^{(0)}$};
    
    \begin{scope}[on background layer]
        \node[draw=orange!60, fill=orange!4, rounded corners=6pt, 
              minimum width=4.8cm, minimum height=3.6cm,
              thick, dashed] (gatebg) at (1.3, -0.1) {};
    \end{scope}
    \node[font=\footnotesize\bfseries, text=orange!70!black] at (1.3, 1.4) {Variability Gating Layer};
    \node[annot] at (1.3, 1.05) {(${\sim}$1K params, $<$1\% overhead)};
    
    \node[featurebox] (region) at (0.15, 0.65) {Region Tag\\[-2pt]$r_i \!\in\! \{0,1\}$};
    \node[featurebox] (pos) at (1.35, 0.65) {Position\\[-2pt]PE$(i)$};
    \node[featurebox] (gc) at (2.55, 0.65) {GC Content\\[-2pt]$\text{GC}_i$};
    
    \node[featurebox] (entropy) at (0.15, -0.2) {Entropy\\[-2pt]$H_i$};
    
    \node[gatebox, minimum width=1.8cm] (concat) at (1.95, -0.2) {$\vf_i$};
    
    \draw[grayarrow] (region.south) -- ++(0,-0.12) -| ($(concat.north)+(-0.6,0)$);
    \draw[grayarrow] (pos.south) -- ($(concat.north)+(0,0)$);
    \draw[grayarrow] (gc.south) -- ++(0,-0.12) -| ($(concat.north)+(0.6,0)$);
    \draw[grayarrow] (entropy.east) -- (concat.west);
    
    \node[gatebox] (gate) at (1.95, -1.0) {Gate $g_i = \sigma(\cdot)$};
    \draw[arrow] (concat) -- (gate);
    
    \draw[arrow] (embed.east) -- (-1.2, 0) -- (-1.2, -1.0) -- (-0.4, -1.0);
    
    \node[draw, circle, minimum size=0.42cm, fill=white, inner sep=0pt, font=\small] (mult) at (-0.4, -1.0) {$\odot$};
    \draw[arrow] (gate.west) -- (mult.east);
    
    \node[eqlabel] at (0.75, -1.55) {$\tilde{\vh}_i^{(0)} = g_i \odot \vh_i^{(0)}$};
    
    \node[mainbox, minimum width=1.8cm, minimum height=1.0cm] (transformer) at (5.0, -1.0) {Transformer\\[-2pt]Encoder};
    \draw[arrow] (3.6, -1.0) -- (transformer.west);
    \node[annot] at (5.0, -1.8) {$N{=}6$ layers, $d{=}256$, 8 heads};
    
    \node[box, fill=gray!10, minimum width=1.0cm] (cls) at (7.0, -1.0) {[CLS]};
    \draw[arrow] (transformer.east) -- (cls.west);
    \node[eqlabel] at (7.0, -1.5) {$\vh_{\text{CLS}}$};
    
    \node[headbox, fill=purple!12] (media) at (9.0, 0.1) {Media (42)};
    \node[headbox, fill=teal!12] (temp) at (9.0, -0.55) {Temp (3)};
    \node[headbox, fill=cyan!12] (ph) at (9.0, -1.2) {pH (3)};
    \node[headbox, fill=lime!15] (o2) at (9.0, -1.85) {O$_2$ (4)};
    
    \draw[arrow] (cls.east) -- ++(0.3, 0) |- (media.west);
    \draw[arrow] (cls.east) -- ++(0.3, 0) |- (temp.west);
    \draw[arrow] (cls.east) -- ++(0.3, 0) |- (ph.west);
    \draw[arrow] (cls.east) -- ++(0.3, 0) |- (o2.west);
    
    \node[label] at (9.0, 0.65) {Predictions};
    
    \draw[arrow] (media.east) -- ++(0.35, 0);
    \draw[arrow] (temp.east) -- ++(0.35, 0);
    \draw[arrow] (ph.east) -- ++(0.35, 0);
    \draw[arrow] (o2.east) -- ++(0.35, 0);
    
    \end{tikzpicture}%
    }
    \caption{Architecture of the variability-gated encoder. The input 16S rRNA sequence is embedded with positional encodings, then modulated by a lightweight gating layer (dashed box) that computes position-wise importance weights. The gate conditions on biological features: region tags indicating hypervariable (V1--V9) vs.\ conserved positions, positional encoding, local GC content, and Shannon entropy. Gated embeddings are processed by a Transformer encoder, and a [CLS] token representation feeds four task-specific heads for media, temperature, pH, and oxygen requirement predictions.}
    \label{fig:framework}
\end{figure*}

\subsection{Architecture Overview}
\label{sec:architecture}

Our model consists of four main components (\Cref{fig:framework}):

\textbf{Embedding Layer.}
Nucleotides are embedded into $d$ dimensions with sinusoidal positional encodings: $\mathbf{h}_i^{(0)} = \text{Embed}(x_i) + \text{PE}(i) \in \mathbb{R}^d$.
where $x_i$ denotes the nucleotide at position $i$ and $\text{PE}(i)$ denotes the sinusoidal positional encoding.

\textbf{Variability Gating Layer.}
To encode the known structure of 16S sequences into the learned representations, we apply gating before the main encoder. This design allows the model to allocate representational capacity within embeddings rather than modifying attention patterns post-hoc. The gating mechanism modulates each position's representation based on biological features:
\begin{equation}
    \tilde{\mathbf{h}}_i^{(0)} = g_i \odot \mathbf{h}_i^{(0)}
\end{equation}
where $g_i \in \mathbb{R}^d$ is a learned gate vector and $\odot$ denotes element-wise multiplication.

\textbf{Transformer Encoder.}
The gated embeddings are processed by a standard Transformer encoder with $N$ layers of multi-head self-attention and feed-forward networks:
\begin{equation}
    \mathbf{H}^{(N)} = \text{TransformerEncoder}(\tilde{\mathbf{H}}^{(0)})
\end{equation}

\textbf{Task-Specific Heads.}
For each task $t$, a separate linear head predicts
$\hat{y}^{(t)} = \mathrm{softmax}(\mathbf{W}^{(t)} \mathbf{h}_{\mathrm{CLS}} + \mathbf{b}^{(t)})$,
where $\mathbf{h}_{\mathrm{CLS}}$ is the final [CLS] embedding.

\subsection{Gating Mechanism}
\label{sec:gating}

The gating layer is the core contribution of our architecture. It computes position-wise importance weights conditioned on three types of features:

\textbf{Region Tags.}
We annotate each sequence position with a binary indicator distinguishing between hypervariable regions (V1--V9) and conserved regions. To determine region boundaries, sequences are aligned to the SILVA SSU Ref NR 138.1 database~\citep{quastSILVARibosomalRNA2012} using the SINA aligner (v1.7.2). This process provides  
position-wise mappings to the reference alignment where V-region boundaries are pre-annotated. Formally, let $r_i \in \{0, 1\}$ indicate the hypervariable status for position $i$.

\textbf{Local Sequence Complexity.}
We compute two local complexity measures in a sliding window of size $w$ centered at each position:
\begin{itemize}[nosep,leftmargin=*]
    \item \textbf{GC content}: $\text{GC}_i = \frac{1}{w} \sum_{j=i-w/2}^{i+w/2} \mathbf{1}[x_j \in \{G, C\}]$, where $\mathbf{1}[\cdot]$ is the indicator function.
    \item \textbf{Shannon entropy}: $H_i = -\sum p_b^{(i)} \log p_b^{(i)}$, where $p_b^{(i)}$ is the local frequency of base $b$.
\end{itemize}

\textbf{Gate Computation.}
The gate vector at each position is computed by a single linear layer:
\begin{equation}
    \mathbf{f}_i = [r_i;\, p_i;\, \text{GC}_i;\, H_i] \in \mathbb{R}^{4}
\end{equation}
where $r_i \in \{0, 1\}$ is the region tag, $p_i = i/L \in [0, 1]$ is the normalized position (with $L$ the sequence length), and $\text{GC}_i, H_i$ are local complexity measures. The gate is then:
\begin{equation}
    g_i = \sigma(\mathbf{W}_g \mathbf{f}_i + \mathbf{b}_g) \in \mathbb{R}^d
    \label{eq:gate}
\end{equation}
where $\sigma$ is the sigmoid function, ensuring gate values lie in $[0, 1]$, and $\mathbf{W}_g \in \mathbb{R}^{d \times 4}$, $\mathbf{b}_g \in \mathbb{R}^d$. The gate operates token-wise: each position receives an independent gate vector computed from its local features. With $d=256$, the gate network adds $4 \times 256 + 256 = 1{,}280$ parameters ($<$1\% of total model size). We intentionally keep the module low-capacity to prevent overfitting, thereby forcing the model to leverage the provided biological features rather than learning arbitrary gating patterns.

\textbf{Interpretation.}
Functionally, a gate value approaching 1 preserves the original embedding, while values near 0 suppress the representation. By conditioning on region tags and complexity features, the model can learn which positions carry informative signal for cultivation prediction. Importantly, this constitutes a soft inductive bias: the model is encouraged but not forced to weight regions differently, and the final gating pattern is learned from data.

\subsection{Training Objective}
\label{sec:training}

We train the model using a multi-task objective that combines cross-entropy losses for each prediction task, $\mathcal{L} = \sum_{t} \lambda_t \mathcal{L}_{\text{CE}}^{(t)},$
where for task $t$ the model outputs logits $\mathbf z^{(t)} \in \mathbb{R}^{C_t}$,
$\hat{\mathbf y}^{(t)}=\mathrm{softmax}(\mathbf z^{(t)})$ denotes the predicted probability vector,
and $\mathbf y^{(t)} \in \{0,1\}^{C_t}$ is the one-hot ground-truth label.
The per-task cross-entropy loss is $\mathcal{L}_{\text{CE}}^{(t)} = -\sum_{c=1}^{C_t} y^{(t)}_c \log \hat{y}^{(t)}_c,$
and $\lambda_t$ are task-specific weights.

\textbf{Loss Weighting.}
To handle class imbalance, we use inverse-frequency class weights computed on the training split. For task $t$ and class $c$ with count $n_c^{(t)}$, we set $w_c^{(t)} \propto 1/(n_c^{(t)}+\epsilon)$ and apply $w_{y}^{(t)}$ to the cross-entropy term. Task weights $\lambda_t$ balance contributions across tasks with numbers of classes.

\textbf{Regularization.}
We apply dropout to the Transformer layers and add an L2 penalty on the gate network parameters to encourage sparse, interpretable gating patterns.

\section{Experiments}
\label{sec:experiments}

\subsection{Baselines and Implementation}
\label{sec:impl_details}

We benchmark against a diverse set of methods spanning traditional ML, similarity-based retrieval, deep learning, and pretrained models. To isolate the gating effect, all neural methods share identical classification heads, losses, and early stopping criteria, with only the encoder changed. Models trained from scratch use 100 epochs, while RNABERT fine-tuning uses 50 epochs due to faster convergence.

\textbf{Traditional ML.}
\textbf{$k$-mer + RF/XGBoost}: Extract $k$-mer frequencies ($k \in \{3,4,5\}$, 1,344 features) and train Random Forest \citep{Breiman:2001hzm} or XGBoost \citep{Chen:2016btl} classifiers independently for each task. Tests whether compositional features suffice.

\textbf{Similarity-Based.}
\textbf{GROWREC} \citep{oberhardtHarnessingLandscapeMicrobial2015}: Predicts by majority vote among $k=5$ phylogenetically nearest neighbors. \textbf{$k$-NN (Sequence)}: Predicts by majority vote among $k=5$ nearest neighbors using normalized Levenshtein distance to account for variable sequence lengths.

\textbf{Deep Learning.}
\textbf{CNN}: Stacked convolutions with global pooling \citep{Lecun:1998iof}; captures local motifs but limited long-range dependencies. \textbf{Transformer (Flat)}: Standard encoder \citep{vaswaniAttentionAllYou2017} with learned positional embeddings; serves as the ``no inductive bias'' baseline.

\textbf{Pretrained.}
\textbf{RNABERT} \citep{akiyamaInformativeRNABase2022}: Fine-tuned pretrained RNA model.

\textbf{Implementation.}
Our Transformer uses $N=6$ layers, $d=256$, 8 attention heads, and a $1{,}280$-parameter gate (4-dim biological features $\to$ 256-dim gate; $w=21$ complexity window). For RNABERT, we fine-tune the public 440-token checkpoint with center-cropped inputs and randomly initialized task heads. All models train with AdamW + linear warmup, select the checkpoint with best validation joint accuracy, and report mean over 5 seeds. Full hyperparameters in Appendix \ref{app:implementation}.

\subsection{Main Results}
\label{sec:main_results}

\begin{table*}[!t]
    \caption{Main results on the DSNI test set. Best in \textbf{bold}, second-best \underline{underlined}; mean $\pm$ std over 5 seeds. Params $=$ trainable parameters in millions.}
    \label{tab:main_results}
    \centering
    \begin{small}
    \begin{tabular}{lcccccc}
        \toprule
        \textbf{Method} & \textbf{Params} & \textbf{Media Acc.} & \textbf{Temp. Acc.} & \textbf{pH Acc.} & \textbf{O$_2$ Acc.} & \textbf{Joint Acc.} \\
        \midrule
        \multicolumn{7}{l}{\textit{Traditional ML}} \\
        $k$-mer + RF & --- & 38.2{\tiny$\pm$1.1} & 86.4{\tiny$\pm$0.8} & 89.1{\tiny$\pm$0.7} & 72.3{\tiny$\pm$0.9} & 24.7{\tiny$\pm$1.2} \\
        $k$-mer + XGBoost & --- & 41.5{\tiny$\pm$0.9} & 87.8{\tiny$\pm$0.6} & 90.3{\tiny$\pm$0.5} & 74.6{\tiny$\pm$0.8} & 27.3{\tiny$\pm$1.0} \\
        \midrule
        \multicolumn{7}{l}{\textit{Phylogenetic / Similarity-Based}} \\
        GROWREC & --- & 46.3{\tiny$\pm$0.0} & 85.2{\tiny$\pm$0.0} & 88.4{\tiny$\pm$0.0} & 71.8{\tiny$\pm$0.0} & 28.9{\tiny$\pm$0.0} \\
        $k$-NN (Sequence) & --- & 43.8{\tiny$\pm$0.0} & 87.1{\tiny$\pm$0.0} & 89.7{\tiny$\pm$0.0} & 73.5{\tiny$\pm$0.0} & 27.6{\tiny$\pm$0.0} \\
        \midrule
        \multicolumn{7}{l}{\textit{Deep Learning (No Pretraining)}} \\
        CNN Encoder & 2.1M & 44.7{\tiny$\pm$1.3} & 88.5{\tiny$\pm$0.7} & 90.8{\tiny$\pm$0.6} & 76.2{\tiny$\pm$1.1} & 31.2{\tiny$\pm$1.4} \\
        Transformer (Flat) & 4.8M & 48.3{\tiny$\pm$0.8} & 89.7{\tiny$\pm$0.5} & 91.5{\tiny$\pm$0.4} & 78.4{\tiny$\pm$0.7} & 36.8{\tiny$\pm$0.9} \\
        \textbf{Transformer + Gating} & 4.8M & \underline{52.6}{\tiny$\pm$0.7} & \underline{91.2}{\tiny$\pm$0.4} & \underline{92.8}{\tiny$\pm$0.3} & \underline{81.3}{\tiny$\pm$0.6} & \underline{41.5}{\tiny$\pm$0.8} \\
        \midrule
        \multicolumn{7}{l}{\textit{Pretrained Models}} \\
        RNABERT & 120M & 51.4{\tiny$\pm$0.6} & 90.8{\tiny$\pm$0.4} & 92.4{\tiny$\pm$0.3} & 80.6{\tiny$\pm$0.5} & 39.7{\tiny$\pm$0.7} \\
        \textbf{RNABERT + Gating} & 120M & \textbf{54.8}{\tiny$\pm$0.5} & \textbf{92.1}{\tiny$\pm$0.3} & \textbf{93.5}{\tiny$\pm$0.3} & \textbf{82.7}{\tiny$\pm$0.4} & \textbf{44.2}{\tiny$\pm$0.6} \\
        \bottomrule
    \end{tabular}
    \end{small}
\end{table*}

\Cref{tab:main_results} presents the main benchmark results on DSNI. We report mean $\pm$ std over 5 random seeds. Paired bootstrap confidence intervals with 10{,}000 resamples and McNemar's tests on prediction-level outcomes (reported in Appendix~\ref{app:significance}) show that all gating-induced gains are statistically significant ($p < 0.001$). The results support five observations.

(1) \textbf{Learned representations outperform phylogenetic matching}: Transformer+Gating exceeds GROWREC by +6.3\% media accuracy and +12.6\% joint accuracy. 

(2) \textbf{Gating improves flat encoders}: +4.3\% media, +2.9\% O$_2$; region tags alone contribute +3.3\%. 

(3) \textbf{Inductive bias and pretraining are complementary}: While Transformer+Gating (41.5\%) outperforms RNABERT (39.7\%), the combined RNABERT+Gating yields the best results (44.2\%). 

(4) \textbf{Media Prediction is the most challenging task}: 42 classes with multiple valid answers. Top-3 accuracy (crediting any of the top-3 predictions matching ground truth) reaches 71.2\% for Transformer+Gating, 68.4\% for flat Transformer, and 74.8\% for RNABERT+Gating.

(5) \textbf{Inductive bias improves sample efficiency}: the gating advantage is largest at low data (+8.4\% at 5\% training data; \Cref{sec:few_shot}), suggesting that explicit bias complements data-driven learning.

\textbf{Joint Accuracy in Practical Context.}
While the strict 44.2\% exact-match joint accuracy may appear modest, it underestimates practical utility: 76.3\% of test strains have at least 3 of 4 conditions correct (4/4: 44.2\%; 3/4: 32.1\%), and 93.1\% have at least 2 of 4 correct (\Cref{app:joint_breakdown}). This indicates that many errors are partial rather than completely incorrect protocols. Per-task top-1 accuracies further show that media, with 42 long-tailed classes and multiple plausible alternatives, is the dominant bottleneck, while temperature, pH, and oxygen each exceed 82\% top-1.

\subsection{Low-Data Performance}
\label{sec:few_shot}

\begin{figure}[!htbp]
    \centering
    \begin{tikzpicture}[
        scale=0.9,
        every node/.style={font=\small},
    ]
    
    \draw[->] (0, 0) -- (6.5, 0) node[right, font=\small] {\%};
    \draw[->] (0, 0) -- (0, 5.2) node[above, anchor=south, font=\small, yshift=2mm] {Joint Accuracy (\%)};
    
    \foreach \x/\lab in {0.6/5, 1.5/10, 3/25, 4.5/50, 6/100} {
        \draw (\x, -0.1) -- (\x, 0.1);
        \node[below, font=\footnotesize] at (\x, -0.15) {\lab};
    }
    
    \foreach \y/\lab in {1/10, 2/20, 3/30, 4/40, 5/50} {
        \draw (-0.1, \y) -- (0.1, \y);
        \node[left, font=\footnotesize] at (-0.15, \y) {\lab};
    }
    
    \foreach \y in {1, 2, 3, 4, 5} {
        \draw[gray!30, thin] (0.1, \y) -- (6.3, \y);
    }
    
    
    \fill[blue!20, opacity=0.5] 
        (0.6, 1.69) -- (1.5, 2.34) -- (3, 3.04) -- (4.5, 3.39) -- (6, 3.59)
        -- (6, 3.77) -- (4.5, 3.57) -- (3, 3.24) -- (1.5, 2.58) -- (0.6, 1.97)
        -- cycle;
    
    \fill[orange!25, opacity=0.5] 
        (0.6, 2.55) -- (1.5, 3.02) -- (3, 3.59) -- (4.5, 3.87) -- (6, 4.07)
        -- (6, 4.23) -- (4.5, 4.03) -- (3, 3.77) -- (1.5, 3.22) -- (0.6, 2.79)
        -- cycle;
    
    \fill[green!20, opacity=0.5] 
        (0.6, 2.75) -- (1.5, 3.20) -- (3, 3.55) -- (4.5, 3.74) -- (6, 3.90)
        -- (6, 4.04) -- (4.5, 3.88) -- (3, 3.69) -- (1.5, 3.36) -- (0.6, 2.93)
        -- cycle;
    
    \fill[red!20, opacity=0.5] 
        (0.6, 3.23) -- (1.5, 3.68) -- (3, 4.02) -- (4.5, 4.20) -- (6, 4.36)
        -- (6, 4.48) -- (4.5, 4.32) -- (3, 4.14) -- (1.5, 3.82) -- (0.6, 3.39)
        -- cycle;
    
    \draw[blue!70, thick, dashed, mark=square*, mark options={solid, fill=blue!70, scale=0.8}] 
        plot coordinates {(0.6, 1.83) (1.5, 2.46) (3, 3.14) (4.5, 3.48) (6, 3.68)};
    
    \draw[orange!80!red, thick, mark=*, mark options={fill=orange!80!red, scale=0.9}] 
        plot coordinates {(0.6, 2.67) (1.5, 3.12) (3, 3.68) (4.5, 3.95) (6, 4.15)};
    
    \draw[green!60!black, thick, dash dot, mark=triangle*, mark options={solid, fill=green!60!black, scale=0.9}] 
        plot coordinates {(0.6, 2.84) (1.5, 3.28) (3, 3.62) (4.5, 3.81) (6, 3.97)};
    
    \draw[red!70!black, thick, mark=diamond*, mark options={fill=red!70!black, scale=0.9}] 
        plot coordinates {(0.6, 3.31) (1.5, 3.75) (3, 4.08) (4.5, 4.26) (6, 4.42)};
    
    \draw[<->, gray, thin] (0.9, 2.12) -- (0.9, 2.67);
    \node[right, font=\tiny, gray] at (0.9, 2.6) {+8.4\%};
    
    \begin{scope}[shift={(3.8, 1)}]
        \fill[white, opacity=0.9] (-0.1, -0.45) rectangle (2.8, 0.85);
        \draw[gray!50, thin] (-0.1, -0.45) rectangle (2.8, 0.85);
        
        \draw[blue!70, thick, dashed] (0, 0.6) -- (0.4, 0.6);
        \node[right, font=\tiny] at (0.45, 0.6) {Transformer (Flat)};
        
        \draw[orange!80!red, thick] (0, 0.3) -- (0.4, 0.3);
        \fill[orange!80!red] (0.2, 0.3) circle (1.2pt);
        \node[right, font=\tiny] at (0.45, 0.3) {+ Gating (Ours)};
        
        \draw[green!60!black, thick, dash dot] (0, 0) -- (0.4, 0);
        \node[right, font=\tiny] at (0.45, 0) {RNABERT};
        
        \draw[red!70!black, thick] (0, -0.3) -- (0.4, -0.3);
        \node[right, font=\tiny] at (0.45, -0.3) {RNABERT + Gating};
    \end{scope}
    
    \end{tikzpicture}
    \caption{Few-shot learning curves for joint accuracy across training set sizes. Shaded regions indicate $\pm$1 standard deviation over 3 random seeds. Gating improves performance most strongly in the low-data regime, and RNABERT+Gating performs best across data sizes.}
    \label{fig:few_shot}
\end{figure}
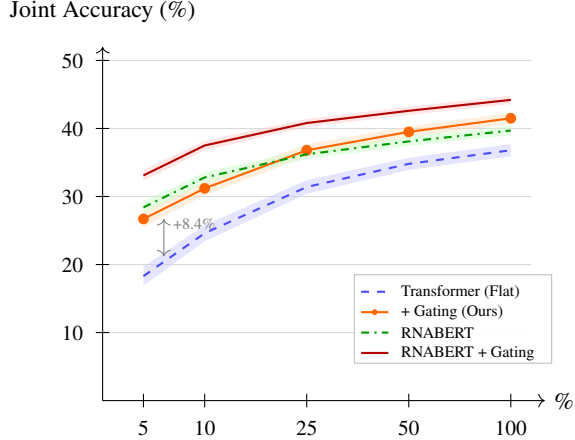

To test whether inductive bias helps in low-data regimes, we train on data subsets (5\%--100\%) and evaluate on the full test set (\Cref{fig:few_shot}). The gating advantage is largest at 5\% data (+8.4\% joint accuracy: 26.7\% vs.\ 18.3\%) and decreases to +4.7\% at full data. RNABERT + Gating achieves 33.1\% at 5\% data, approaching flat Transformer's full-data performance (36.8\%). This pattern suggests that inductive bias is most valuable when data is insufficient for the model to discover relevant patterns independently. As training data increases, the flat encoder gradually exploits sequence structure and narrows the gap, but explicit bias provides a head start that persists even at full data.


\subsection{Robustness and Practical Utility}
\label{sec:robustness_utility}

We further characterize DSNI and our model along three axes: phylogenetic leakage, generalization across phylogenetic distance, and the model's utility as a search-space reducer for wet-lab cultivation.

\textbf{Quantifying Phylogenetic Leakage.}
To measure how much apparent predictability comes from phylogenetic memorization, we re-evaluated three encoder families under random splits and our genus-aware splits while holding all other settings fixed. Joint accuracy drops by $20.1$--$21.4$\,pp across all three encoders (See Appendix \ref{app:leakage}). This substantial drop shows that random splits can strongly inflate cultivation-prediction performance and motivates phylogeny-aware evaluation.

\textbf{Generalization vs.\ Phylogenetic Distance.}
We stratify the test set by 16S identity to the nearest training-set genus. The advantage of RNABERT+Gating over the phylogenetic-matching baseline GROWREC widens with distance, from $+13.7$\,pp at $>$90\% identity to $+16.8$\,pp at $<$80\% (See Appendix~\ref{app:phylo_distance}). This pattern is inconsistent with pure nearest-neighbor matching and suggests that the encoder learns sequence-level signals that remain useful when close training relatives are unavailable.

\textbf{Top-$k$ Recall as a Screening Tool.}
Microbiologists typically screen a shortlist rather than commit to a single protocol, so we also evaluate the model as a search-space reducer (\Cref{tab:topk_recall}). Testing the top-25 ranked joint protocols recovers $90.2$\% of ground-truth protocols, compared to $1{,}512 = 42 \times 3 \times 3 \times 4$ exhaustive combinations. This reduces the candidate set by $\sim\!60\times$ while retaining high recall.

\textbf{Functional-Equivalence Audit of Media Errors.}
A manual audit of 200 top-1 media errors finds that 58\% occur between functionally related media, as detailed in Appendix~\ref{app:media_audit}. This suggests that exact-match media accuracy may underestimate practical recommendation quality.

\subsection{Ablation Studies}
\label{sec:ablations}

We ablate the gating components to quantify their individual contributions (\Cref{tab:ablations}).

\begin{table}[!t]
\centering
\begin{minipage}[b]{0.46\textwidth}
\centering
\begin{small}
\begin{tabular}{lc}
    \toprule
    \textbf{Top-$k$} & \textbf{Joint recall} \\
    \midrule
    $k=1$ & 44.2\% \\
    $k=5$ & 67.8\% \\
    $k=10$ & 78.3\% \\
    $k=20$ & 87.1\% \\
    $k=25$ & 90.2\% \\
    Exhaustive ($1{,}512$) & 100\% \\
    \bottomrule
\end{tabular}
\end{small}
\captionof{table}{Joint protocol recall vs.\ shortlist size (RNABERT+Gating). Exhaustive screening requires enumerating $1{,}512$ protocols.}
\label{tab:topk_recall}
\end{minipage}\hfill
\begin{minipage}[b]{0.50\textwidth}
\centering
\begin{small}
\begin{tabular}{lc}
    \toprule
    \textbf{Configuration} & \textbf{Joint Acc.} \\
    \midrule
    Full model (all features) & \textbf{41.5}{\tiny$\pm$0.8} \\
    \midrule
    \quad $-$ Region tags & 38.2{\tiny$\pm$0.9} \\
    \quad $-$ Position encoding & 40.1{\tiny$\pm$0.8} \\
    \quad $-$ GC content & 40.8{\tiny$\pm$0.8} \\
    \quad $-$ Shannon entropy & 41.0{\tiny$\pm$0.8} \\
    \midrule
    Learned gating (no bio.\ features) & 38.6{\tiny$\pm$0.9} \\
    No gating (Flat Transformer) & 36.8{\tiny$\pm$0.9} \\
    \bottomrule
\end{tabular}
\end{small}
\captionof{table}{Ablation on gating components: joint accuracy on the DSNI test set.}
\label{tab:ablations}
\end{minipage}
\end{table}

\textbf{Feature Contributions.}
Removing region tags causes the largest drop ($-$3.3\%), indicating that the hypervariable/conserved distinction is the most informative gating feature.
Position contributes $-$1.4\%, while complexity features contribute $-$0.7\% (GC) and $-$0.5\% (Shannon).
The ``Learned gating (no bio.\ features)'' configuration replaces the biological feature vector with a learned lookup table over 32 position bins, allowing the model to learn position-dependent gating without explicit biological annotations. 
It achieves 38.6\%, better than no gating (36.8\%) but worse than full model (41.5\%), showing that biologically informed conditioning provides more than a gating regularizer.

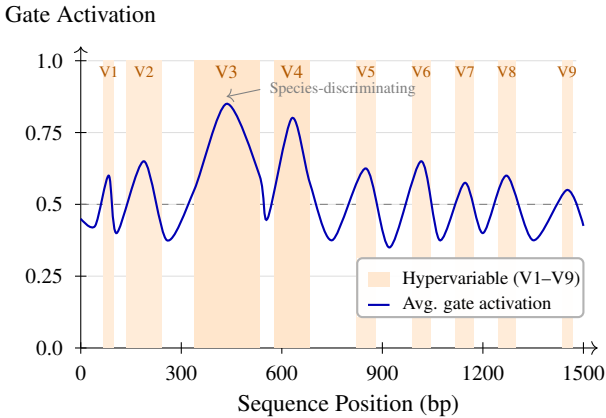
\begin{figure}[!htbp]
    \centering
    \begin{tikzpicture}[
        scale=0.95,
        every node/.style={font=\small},
    ]
    
    \draw[->] (0, 0) -- (7.2, 0) node[below, font=\small, xshift=-3.3cm, yshift=-0.5cm] {Sequence Position (bp)};
    \draw[->] (0, 0) -- (0, 4.2) node[above, anchor=south, font=\small, yshift=2mm] {Gate Activation};
    
    \foreach \x/\lab in {0/0, 1.4/300, 2.8/600, 4.2/900, 5.6/1200, 7/1500} {
        \draw (\x, -0.08) -- (\x, 0.08);
        \node[below, font=\footnotesize] at (\x, -0.12) {\lab};
    }
    
    \foreach \y/\lab in {0/0.0, 1/0.25, 2/0.5, 3/0.75, 4/1.0} {
        \draw (-0.08, \y) -- (0.08, \y);
        \node[left, font=\footnotesize] at (-0.12, \y) {\lab};
    }
    
    \foreach \y in {1, 2, 3, 4} {
        \draw[gray!25, thin] (0.1, \y) -- (6.9, \y);
    }
    
    \draw[gray, dashed, thin] (0, 2) -- (7, 2);
    
    
    \fill[orange!15] (0.32, 0) rectangle (0.46, 4);   
    \fill[orange!15] (0.64, 0) rectangle (1.13, 4);   
    \fill[orange!20] (1.58, 0) rectangle (2.49, 4);   
    \fill[orange!20] (2.69, 0) rectangle (3.19, 4);   
    \fill[orange!15] (3.84, 0) rectangle (4.11, 4);   
    \fill[orange!15] (4.61, 0) rectangle (4.87, 4);   
    \fill[orange!15] (5.22, 0) rectangle (5.48, 4);   
    \fill[orange!15] (5.81, 0) rectangle (6.05, 4);   
    \fill[orange!15] (6.70, 0) rectangle (6.85, 4);   
    
    \node[font=\tiny, orange!70!black] at (0.39, 3.85) {V1};
    \node[font=\tiny, orange!70!black] at (0.88, 3.85) {V2};
    \node[font=\scriptsize, orange!70!black] at (2.03, 3.85) {V3};
    \node[font=\scriptsize, orange!70!black] at (2.94, 3.85) {V4};
    \node[font=\tiny, orange!70!black] at (3.97, 3.85) {V5};
    \node[font=\tiny, orange!70!black] at (4.74, 3.85) {V6};
    \node[font=\tiny, orange!70!black] at (5.35, 3.85) {V7};
    \node[font=\tiny, orange!70!black] at (5.93, 3.85) {V8};
    \node[font=\tiny, orange!70!black] at (6.77, 3.85) {V9};
    
    \draw[blue!70!black, thick, smooth] plot coordinates {
        (0, 1.8)      
        (0.2, 1.7)    
        (0.39, 2.4)   
        (0.5, 1.6)    
        (0.88, 2.6)   
        (1.2, 1.5)    
        (1.58, 2.2)   
        (2.03, 3.4)   
        (2.49, 2.4)   
        (2.6, 1.8)    
        (2.94, 3.2)   
        (3.19, 2.3)   
        (3.5, 1.5)    
        (3.97, 2.5)   
        (4.3, 1.4)    
        (4.74, 2.6)   
        (5.0, 1.5)    
        (5.35, 2.3)   
        (5.6, 1.6)    
        (5.93, 2.4)   
        (6.3, 1.5)    
        (6.77, 2.2)   
        (7.0, 1.7)    
    };
    
    \draw[<-, thin, gray] (2.03, 3.5) -- (2.5, 3.6) node[right, font=\tiny] {Species-discriminating};
    
    \begin{scope}[shift={(4, 0.6)}]
        \fill[white, opacity=0.95] (-0.15, -0.17) rectangle (3, 0.65);
        \draw[gray!60, thick, rounded corners=2pt] (-0.15, -0.17) rectangle (3, 0.65);
        \fill[orange!18] (0, 0.25) rectangle (0.3, 0.45);
        \node[right, font=\scriptsize] at (0.35, 0.35) {Hypervariable (V1--V9)};
        \draw[blue!70!black, thick] (0, 0) -- (0.3, 0);
        \node[right, font=\scriptsize] at (0.35, 0) {Avg. gate activation};
    \end{scope}
    
    \end{tikzpicture}
    \caption{Average gate activation across 16S rRNA sequence positions. Shaded regions indicate hypervariable regions (V1--V9); the model learns to assign higher gate values to these regions, particularly V3 and V4 which provide the finest species-level discrimination. Conserved regions receive lower weights (near the 0.5 baseline), consistent with their limited discriminative power for cultivation prediction.}
    \label{fig:gate_viz}
\end{figure}

\textbf{Learned Gating Patterns.}
\Cref{fig:gate_viz} visualizes average gate activation  across 16S positions. For each position $i$, we average gate value $\bar{g}_i = \frac{1}{d} \sum_{j=1}^{d} [g_i]_j$ across embedding dimensions and test sequences aligned to the same reference position. Because the gate bias is initialized at $b_g=0$ with small random weights, initial activations are near 0.5 (sigmoid of zero), so deviations reflect learned position-dependent modulation.
The learned profile assigns higher weights to hypervariable regions, especially V3 and V4, which are known to provide strong species-level discrimination \citep{chakravortyDetailedAnalysis16S2007}. 
Conserved regions receive lower gate values, consistent with their more limited usefulness for recommending cultivation conditions.


\textbf{Truncation Control for the RNABERT Comparison.}
To rule out that the gain from adding gating to RNABERT is an artifact of RNABERT's 440-token input cap, we retrain the from-scratch encoders under the same center crop (Appendix \ref{app:truncation}). Center-cropping costs $\sim$3.4\,pp of joint accuracy on the from-scratch encoders, but under matched inputs, gating still improves both the flat Transformer ($+4.8$\,pp) and RNABERT ($+4.5$\,pp). Thus, the gain is not driven by input length.

\section{Discussion}
\label{sec:discussion}

\textbf{Limitations.}
Our work learns correlations useful for recommendation, not biological mechanisms. Limitations include label noise in source databases, information loss from media standardization, and dataset bias toward easily cultured organisms, whereas the uncultured majority spans diverse habitats from soil to extreme environments \citep{fiererMetagenomicSmallSubunitRRNA2007}. The learned gating patterns (preference for V3/V4) align with their utility for species discrimination but reflect what the model finds useful, not causality.

\textbf{Practical Workflow.}
Our model should not be viewed as a single-shot cultivation oracle. Its value lies in narrowing the experimental search space. A tiered workflow could use confidence-thresholded predictions (joint score $>$0.35: 62.8\% accuracy on 41\% of strains) for direct prescription and top-$k$ screening for the remainder, meaningfully reducing the dominant cost in cultivating novel organisms~\citep{lagierCulturingHumanMicrobiota2018}.

\textbf{Future Directions.}
DSNI enables research in multi-modal integration (e.g., combining 16S with genome features for the strain subset where genomes are available), uncertainty quantification for calibrated confidence under top-$k$ screening, active learning to feed wet-lab outcomes back into the model, and transfer to other culture collections.

\section{Conclusion}
\label{sec:conclusion}

We introduced DSNI, a 65{,}023-strain 16S benchmark with phylogeny-aware splits, and a Variability-Gated Encoder whose domain-informed inductive bias complements large-scale pretraining. Beyond the headline numbers, the combined system narrows the cultivation search space by $\sim$60$\times$ through top-$k$ screening, providing a tangible step toward sequence-guided cultivation of the uncultured majority. The dataset, code, and trained models are available at \url{https://anonymous.4open.science/r/3B8F}.

\bibliographystyle{plainnat}
\bibliography{ref}

\newpage
\appendix
\section{Additional Dataset Details}
\label{app:dataset}

\subsection{Database Schema}
\label{app:dataset_schema}
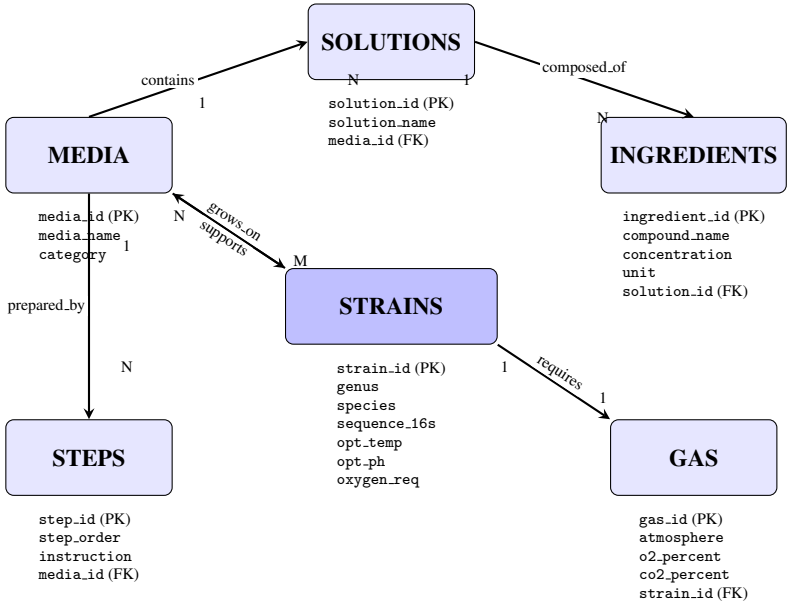
\begin{figure}[!htbp]
    \centering
    \begin{tikzpicture}[
        entity/.style={draw, rectangle, rounded corners=3pt, minimum width=2.2cm, minimum height=1cm, fill=blue!10, font=\small\bfseries, align=center},
        attribute/.style={font=\tiny, align=left},
        relationship/.style={->, thick, >=stealth},
        label/.style={font=\tiny, fill=white, inner sep=1pt},
    ]
    
    \node[entity, fill=blue!25, minimum width=2.8cm] (strains) at (0, 0) {STRAINS};
    \node[attribute, below=0.1cm of strains, anchor=north] {
        \texttt{strain\_id} (PK)\\
        \texttt{genus}\\
        \texttt{species}\\
        \texttt{sequence\_16s}\\
        \texttt{opt\_temp}\\
        \texttt{opt\_ph}\\
        \texttt{oxygen\_req}
    };
    
    \node[entity] (media) at (-4, 2) {MEDIA};
    \node[attribute, below=0.1cm of media, anchor=north] {
        \texttt{media\_id} (PK)\\
        \texttt{media\_name}\\
        \texttt{category}
    };
    
    \node[entity] (solutions) at (0, 3.5) {SOLUTIONS};
    \node[attribute, below=0.1cm of solutions, anchor=north] {
        \texttt{solution\_id} (PK)\\
        \texttt{solution\_name}\\
        \texttt{media\_id} (FK)
    };
    
    \node[entity] (ingredients) at (4, 2) {INGREDIENTS};
    \node[attribute, below=0.1cm of ingredients, anchor=north] {
        \texttt{ingredient\_id} (PK)\\
        \texttt{compound\_name}\\
        \texttt{concentration}\\
        \texttt{unit}\\
        \texttt{solution\_id} (FK)
    };
    
    \node[entity] (steps) at (-4, -2) {STEPS};
    \node[attribute, below=0.1cm of steps, anchor=north] {
        \texttt{step\_id} (PK)\\
        \texttt{step\_order}\\
        \texttt{instruction}\\
        \texttt{media\_id} (FK)
    };
    
    \node[entity] (gas) at (4, -2) {GAS};
    \node[attribute, below=0.1cm of gas, anchor=north] {
        \texttt{gas\_id} (PK)\\
        \texttt{atmosphere}\\
        \texttt{o2\_percent}\\
        \texttt{co2\_percent}\\
        \texttt{strain\_id} (FK)
    };
    
    \draw[relationship] (strains.north west) -- node[label, above, sloped] {grows\_on} (media.south east);
    \draw[relationship] (media.south east) -- node[label, below, sloped] {supports} (strains.north west);
    
    \draw[relationship] (media.north) -- node[label, left] {contains} (solutions.west);
    
    \draw[relationship] (solutions.east) -- node[label, above] {composed\_of} (ingredients.north);
    
    \draw[relationship] (media.south) -- node[label, left] {prepared\_by} (steps.north);
    
    \draw[relationship] (strains.south east) -- node[label, above, sloped] {requires} (gas.north west);
    
    \node[font=\tiny] at (-2.8, 1.2) {N};
    \node[font=\tiny] at (-1.2, 0.6) {M};
    \node[font=\tiny] at (-2.5, 2.7) {1};
    \node[font=\tiny] at (-0.5, 3.0) {N};
    \node[font=\tiny] at (1.0, 3.0) {1};
    \node[font=\tiny] at (2.8, 2.5) {N};
    \node[font=\tiny] at (-3.5, 0.8) {1};
    \node[font=\tiny] at (-3.5, -0.8) {N};
    \node[font=\tiny] at (1.5, -0.8) {1};
    \node[font=\tiny] at (2.8, -1.2) {1};
    
    \end{tikzpicture}
    \caption{Entity-relationship diagram of the DSNI database schema. The central STRAINS table contains 16S rRNA sequences and cultivation conditions, linked to MEDIA through a many-to-many relationship. Media compositions are decomposed into SOLUTIONS and INGREDIENTS, while preparation protocols are captured in STEPS. Atmospheric requirements are stored in the GAS table.}
    \label{fig:er_diagram}
\end{figure}

The DSNI database comprises six interconnected tables, illustrated in \Cref{fig:er_diagram}.


\begin{itemize}[leftmargin=*]
    \item \textbf{STRAINS}: Core table containing strain identifiers, taxonomic information (genus, species), 16S rRNA sequences, and links to cultivation conditions. This is the primary table from which we derive our dataset.
    \item \textbf{MEDIA}: Defined growth media with unique identifiers and standardized names. We normalize media names across sources to create 42 canonical categories.
    \item \textbf{SOLUTIONS}: Component solutions used in media preparation (e.g., trace element solutions, vitamin mixtures).
    \item \textbf{INGREDIENTS}: Individual chemical compounds with concentrations and units, enabling fine-grained media composition analysis.
    \item \textbf{STEPS}: Preparation instructions for complex media, preserving protocol details for reproducibility.
    \item \textbf{GAS}: Atmospheric requirements (oxygen levels, CO$_2$ supplementation), from which we derive the four oxygen requirement categories.
\end{itemize}

\Cref{tab:schema} provides detailed field descriptions for key tables.

\begin{table}[!htbp]
    \caption{Key fields in the DSNI database schema.}
    \label{tab:schema}
    \centering
    \begin{small}
    \begin{tabular}{llp{5.5cm}}
        \toprule
        \textbf{Table} & \textbf{Field} & \textbf{Description} \\
        \midrule
        STRAINS & strain\_id & Unique identifier (DSM/BEI number) \\
        & genus & Taxonomic genus assignment \\
        & species & Species epithet \\
        & sequence\_16s & Full or partial 16S rRNA sequence \\
        & opt\_temp & Optimal growth temperature ($^\circ$C) \\
        & opt\_ph & Optimal growth pH \\
        & oxygen\_req & Oxygen requirement category \\
        \midrule
        MEDIA & media\_id & Unique media identifier \\
        & media\_name & Standardized media name \\
        & category & One of 42 normalized categories \\
        \midrule
        GAS & atmosphere & Atmospheric composition description \\
        & o2\_percent & Oxygen percentage (0--21\%) \\
        & co2\_percent & CO$_2$ percentage if supplemented \\
        \bottomrule
    \end{tabular}
    \end{small}
\end{table}

\subsection{Comparison with Prior Cultivation-Prediction Resources}
\label{app:benchmark_compare}

\Cref{tab:benchmark_compare} situates DSNI against the two closest publicly accessible resources for cultivation prediction. KOMODO~\citep{oberhardtHarnessingLandscapeMicrobial2015} provides organism-media pairings under random splits but covers media only; MediaDB is a curated catalog of media compositions without train/test infrastructure. DSNI is roughly $3.6\times$ the size of KOMODO, is the first such resource to pair sequences with all four primary cultivation conditions under a single standardized labeling scheme, and is the first to provide phylogeny-aware splits.

\begin{table}[!htbp]
    \caption{Cultivation-prediction benchmarks. DSNI is the largest resource and the only one to provide genus-aware splits and standardized labels for all four primary conditions.}
    \label{tab:benchmark_compare}
    \centering
    \begin{small}
    \begin{tabular}{lccc}
        \toprule
        \textbf{Resource} & \textbf{Organisms} & \textbf{Labels} & \textbf{Split strategy} \\
        \midrule
        KOMODO~\citep{oberhardtHarnessingLandscapeMicrobial2015} & ${\sim}18$K pairs & Media only & Random \\
        MediaDB & ${\sim}800$ orgs.\ & Media composition & None (catalog) \\
        \textbf{DSNI (ours)} & \textbf{65,023 strains} & \textbf{4 conditions} & \textbf{Genus-aware} \\
        \bottomrule
    \end{tabular}
    \end{small}
\end{table}

\subsection{Choice of 16S as Input}
\label{app:why_16s}

Our choice of 16S rRNA as the model input is driven by accessibility rather than a claim of informational superiority over genome-scale features. As of early 2026, NCBI hosts more than $2.5$ million 16S rRNA gene records for bacterial taxa, whereas RefSeq contains roughly $450$K complete prokaryotic genome assemblies (${\sim}18\%$ coverage). For uncultured organisms detected via environmental metagenomics, 16S amplicon sequences are often the only molecular data available. Among the ${\sim}8.2$K DSNI strains for which we could locate matching genome assemblies, genome-feature classifiers in the style of \citet{sauerPredictingOptimalGrowth2019a} achieve only marginally higher temperature accuracy ($+2.1$\,pp) yet are inapplicable to the remaining 86\% of strains. We therefore position DSNI as a foundation benchmark whose 16S input maximizes coverage, while leaving genome-augmented multi-modal extensions to future work.

ML-based recommendation also fills a complementary niche relative to existing wet-lab cultivation strategies (\Cref{tab:methods_compare}): traditional trial-and-error scales poorly with the diversity of the uncultured majority, culturomics requires expensive multi-condition screening campaigns, and dilution-to-extinction targets only a narrow physiological regime. A 16S-conditioned recommender provides instant, organism-agnostic ranked shortlists that complement these approaches by reducing their search space rather than replacing them.

\begin{table}[!htbp]
    \caption{Wet-lab cultivation methods vs.\ DSNI-based ML recommendation. ML provides instant, organism-agnostic ranked shortlists that complement rather than replace wet-lab methods.}
    \label{tab:methods_compare}
    \centering
    \begin{small}
    \setlength{\tabcolsep}{4pt}
\begin{tabular}{lllll}
        \toprule
        \textbf{Method} & \textbf{Input} & \textbf{Throughput} & \textbf{Coverage} & \textbf{Limitation} \\
        \midrule
        Trial-and-error & Isolate & 50--200 cond./org. & Single organism & Slow, expensive \\
        Culturomics~\citep{lagierCulturingHumanMicrobiota2018} & Fecal/env.\ sample & ${\sim}200$ cond./run & Broad, untargeted & 5--30\% success \\
        Dilution-to-extinction & Liquid sample & Low & Specialized & Low-nutrient only \\
        \textbf{DSNI (ML)} & \textbf{16S only} & \textbf{Instant} & \textbf{Any 16S-tagged org.} & \textbf{Needs validation} \\
        \bottomrule
    \end{tabular}
    \end{small}
\end{table}

\subsection{Quality Control and Deduplication Pipeline}
\label{app:qc_dedup}

We retained sequences between 500 and 1{,}500\,bp, capturing partial and near-full-length 16S amplicons while excluding short fragments and potential chimeras. We also removed sequences with $>$1\% ambiguous bases, eliminated exact duplicates, and aligned them to the SILVA reference~\citep{quastSILVARibosomalRNA2012} to verify gene identity. To prevent data leakage, we clustered sequences at 97\% identity, the standard OTU threshold for species-level grouping~\citep{yarzaUnitingClassificationCultured2014}. Within each cluster, we retained the entry with the most complete metadata across media, temperature, pH, and oxygen, breaking ties in favor of DSMZ records.

\section{Implementation Details}
\label{app:implementation}

\subsection{Hyperparameters}

\Cref{tab:hyperparams} summarizes the hyperparameters used for all models in our experiments.

\begin{table}[!htbp]
    \caption{Hyperparameters for all models. Values were selected via grid search on the validation set.}
    \label{tab:hyperparams}
    \centering
    \begin{small}
    \begin{tabular}{lcc}
        \toprule
        \textbf{Hyperparameter} & \textbf{Transformer} & \textbf{RNABERT} \\
        \midrule
        Hidden dimension $d$ & 256 & 768 (fixed) \\
        Number of layers $N$ & 6 & 12 (fixed) \\
        Attention heads & 8 & 12 (fixed) \\
        Feed-forward dim & 1024 & 3072 (fixed) \\
        Dropout & 0.1 & 0.1 \\
        \midrule
        \multicolumn{3}{l}{\textit{Gating Layer}} \\
        Gate input dim $d_f$ & 4 & 4 \\
        Complexity window $w$ & 21 & 21 \\
        Gate dropout & 0.05 & 0.05 \\
        \midrule
        \multicolumn{3}{l}{\textit{Training}} \\
        Learning rate & $1 \times 10^{-4}$ & $2 \times 10^{-5}$ \\
        Weight decay & 0.01 & 0.01 \\
        Batch size & 32 & 16 \\
        Epochs & 100 & 50 \\
        Warmup ratio & 0.1 & 0.1 \\
        \midrule
        \multicolumn{3}{l}{\textit{Loss Weights} $\lambda_t$} \\
        Media & 1.0 & 1.0 \\
        Temperature & 0.5 & 0.5 \\
        pH & 0.5 & 0.5 \\
        Oxygen & 0.75 & 0.75 \\
        \bottomrule
    \end{tabular}
    \end{small}
\end{table}

\subsection{Training Schedule}
\label{app:training_details}

All models use AdamW with linear warmup over the first 10\% of steps. Learning rates are $1 \times 10^{-4}$ for scratch models and $2 \times 10^{-5}$ for RNABERT fine-tuning; weight decay is $0.01$ throughout. Batch size is 32 (16 for RNABERT due to memory). Models trained from scratch use 100 epochs and RNABERT fine-tuning uses 50 epochs due to faster convergence. We select the checkpoint with best validation joint accuracy and report mean over 5 seeds.

\subsection{Computational Resources}

Training is performed on NVIDIA A100 GPUs. The gated Transformer trains in approximately 8 hours; the flat Transformer baseline requires 7 hours. RNABERT fine-tuning takes approximately 12 hours due to the larger model size. Traditional ML baselines (RF, XGBoost) train in under 30 minutes on CPU.

\subsection{Traditional ML Baselines}

For $k$-mer features, we extract frequency vectors for $k \in \{3, 4, 5\}$, yielding $4^3 + 4^4 + 4^5 = 1,344$ dimensional feature vectors. We apply L2 normalization to account for varying sequence lengths.

\textbf{Random Forest:} 500 trees, max depth 30, min samples per leaf 5, class-weighted for imbalance.

\textbf{XGBoost:} 500 boosting rounds, max depth 8, learning rate 0.1, subsample 0.8, colsample by tree 0.8.

\subsection{Gating Mechanism Design Choices}
\label{app:gating_design}

\paragraph{Multiplicative gating vs.\ alternatives.}
The gate computes position-wise importance \emph{before} attention by element-wise scaling of the embedding $\tilde{\mathbf h}_i^{(0)} = g_i \odot \mathbf h_i^{(0)}$. We considered two alternatives that condition on the same biological feature vector $\mathbf f_i$ (region tag, position, GC, entropy):
\begin{itemize}[leftmargin=*]
    \item \textbf{Concatenation}: append $\mathbf f_i$ to the input embedding ($\mathbf h_i^{(0)} \mathbin\Vert \mathbf f_i$). This adds capacity but does not modulate existing representations, leaving attention to discover any position-dependent emphasis from data.
    \item \textbf{Attention bias}: add a learned bias $b(\mathbf f_i, \mathbf f_j)$ to the attention logits. This modulates pairwise interaction scores but cannot suppress a position's embedding before that interaction occurs.
\end{itemize}
Multiplicative gating uniquely combines (i) explicit position-wise suppression/amplification, (ii) integration of the bias \emph{prior} to attention so that all downstream layers operate on the modulated representation, and (iii) low capacity ($1{,}280$ parameters) that forces the module to leverage the supplied features rather than memorize position-conditional patterns. The ablation in \Cref{tab:ablations} provides a controlled probe: ``Learned gating (no bio.\ features)'' replaces $\mathbf f_i$ with a learned 32-bin position embedding while retaining the multiplicative form. It scores $38.6\%$ vs.\ $41.5\%$ for the full model and $36.8\%$ for no gating, showing that the multiplicative mechanism alone contributes ${\sim}1.8$\,pp of regularization but the bulk of the gain ($+2.9$\,pp) comes from the biological conditioning features.

\paragraph{Independent heads vs.\ structured decoder.}
Although cultivation conditions are interdependent (e.g., thermophilic organisms prefer specialized media), our four task-specific heads are conditionally independent given the shared $[\mathrm{CLS}]$ representation. To verify that this design does not discard usable structure, we estimate the empirical mutual information (MI) between every pair of labels on the training split (\Cref{tab:label_mi}). Pairwise MI is at most $0.34$\,bits, corresponding to MI/$H(\cdot)\approx5$--$7\%$ of the entropy of the source label. Inter-task structure is therefore moderate. We choose independent heads for two reasons: (i) \emph{isolability}---per-task accuracy and ablation deltas can be cleanly attributed to the shared encoder rather than to a structured decoder confound; (ii) \emph{label-noise resilience}---given the heterogeneous, partly hand-curated source metadata, cascading dependencies through a structured decoder risk amplifying label noise on rare classes (psychrophiles, alkaliphiles, microaerophiles).

\begin{table}[!htbp]
    \caption{Empirical mutual information $\mathrm{MI}(A;B)$ between cultivation-condition label pairs on the training split. $\mathrm{MI}/H(A)$ measures how much of $A$'s entropy is explained by $B$.}
    \label{tab:label_mi}
    \centering
    \begin{small}
    \begin{tabular}{lccc}
        \toprule
        \textbf{$A \to B$} & \textbf{$H(A)$} & \textbf{$H(A\mid B)$} & \textbf{$\mathrm{MI}(A;B)$} \\
        \midrule
        Media $\to$ Temperature & 4.61 & 4.38 & 0.23 \\
        Media $\to$ Oxygen & 4.61 & 4.27 & 0.34 \\
        Temperature $\to$ pH & 0.61 & 0.57 & 0.04 \\
        \bottomrule
    \end{tabular}
    \end{small}
\end{table}

\section{Additional Experiments}
\label{app:experiments}

\subsection{Per-Class Performance}

\Cref{tab:perclass} reports F1 scores broken down by class for the temperature and pH prediction tasks, highlighting performance on minority classes. Gating delivers its largest gains on rare classes: psychrophiles improve by $+6.4$\,F1, thermophiles by $+6.8$, acidophiles by $+7.6$, and alkaliphiles by $+7.6$ (all relative to the flat Transformer), confirming that the inductive bias is most useful exactly where data-driven discovery is unreliable.

\begin{table}[!htbp]
    \caption{Per-class F1 scores for temperature and pH prediction. Transformer+Gating model shows improved performance on minority classes (psychrophile, thermophile, acidophile, alkaliphile).}
    \label{tab:perclass}
    \centering
    \begin{small}
    \begin{tabular}{lccc}
        \toprule
        & \textbf{Flat Trans.} & \textbf{+Gating} & \textbf{RNABERT+Gating} \\
        \midrule
        \multicolumn{4}{l}{\textit{Temperature}} \\
        Psychrophile (6\%) & 52.3 & 58.7 & 61.2 \\
        Mesophile (89\%) & 94.2 & 95.1 & 95.8 \\
        Thermophile (5\%) & 48.6 & 55.4 & 58.9 \\
        \midrule
        \multicolumn{4}{l}{\textit{pH}} \\
        Acidophile (4\%) & 45.2 & 52.8 & 56.3 \\
        Neutrophile (91\%) & 95.8 & 96.4 & 97.1 \\
        Alkaliphile (5\%) & 41.7 & 49.3 & 53.6 \\
        \bottomrule
    \end{tabular}
    \end{small}
\end{table}

\Cref{tab:perclass_media} drills further into the media task, where the long tail dominates difficulty. Per-class F1 for RNABERT+Gating drops by roughly an order of magnitude between the most common categories ($n>3{,}000$) and the long-tail rare media ($n<50$), explaining why media is the bottleneck of joint accuracy despite competitive top-1 numbers on the other three tasks.

\begin{table}[!htbp]
    \caption{Per-class F1 scores for the media task (RNABERT+Gating), illustrating the long-tail collapse. Sample counts are training-set frequencies.}
    \label{tab:perclass_media}
    \centering
    \begin{small}
    \begin{tabular}{lcc}
        \toprule
        \textbf{Media category} & \textbf{\# samples} & \textbf{F1} \\
        \midrule
        TSA (most common) & 8,247 & 68.4 \\
        BHI & 5,182 & 61.7 \\
        Nutrient Agar & 4,891 & 59.3 \\
        Marine Agar & 3,256 & 52.8 \\
        \midrule
        Acidic mineral (tail) & 47 & 12.3 \\
        Methanogen medium (tail) & 31 & 8.7 \\
        \bottomrule
    \end{tabular}
    \end{small}
\end{table}

\subsection{Macro-F1 Results}

\Cref{tab:macrof1} reports macro-averaged F1 scores across all methods, complementing the accuracy metrics in the main text.

\begin{table}[!htbp]
    \caption{Macro-F1 scores on the DSNI test set. Macro-averaging weights all classes equally, emphasizing performance on rare classes.}
    \label{tab:macrof1}
    \centering
    \begin{small}
    \begin{tabular}{lcccc}
        \toprule
        \textbf{Method} & \textbf{Media} & \textbf{Temp.} & \textbf{pH} & \textbf{O$_2$} \\
        \midrule
        $k$-mer + RF & 28.4 & 62.3 & 58.7 & 54.2 \\
        $k$-mer + XGBoost & 31.2 & 65.1 & 61.4 & 57.8 \\
        CNN Encoder & 34.6 & 67.8 & 63.2 & 60.4 \\
        Transformer (Flat) & 38.2 & 71.5 & 66.8 & 64.1 \\
        \textbf{Transformer + Gating} & \underline{43.7} & \underline{76.2} & \underline{72.4} & \underline{69.8} \\
        RNABERT & 41.8 & 74.6 & 70.9 & 68.2 \\
        \textbf{RNABERT + Gating} & \textbf{46.2} & \textbf{78.4} & \textbf{75.1} & \textbf{72.5} \\
        \bottomrule
    \end{tabular}
    \end{small}
\end{table}

The gating mechanism provides consistent improvements in macro-F1 (+5.5\% on average over flat Transformer), with the largest gains on media prediction where class imbalance is most severe.

\subsection{Generalization vs.\ Phylogenetic Distance}
\label{app:phylo_distance}

\Cref{tab:phylo_distance} reports joint accuracy stratified by 16S identity to the nearest training-set genus. RNABERT+Gating's advantage over GROWREC widens with phylogenetic distance, evidence the encoder learns sequence-level signals beyond nearest-neighbor matching.

\begin{table}[!htbp]
    \caption{Joint accuracy stratified by 16S identity to the nearest training-set genus. RNABERT+Gating outperforms the phylogenetic-matching baseline GROWREC by an increasing margin at greater phylogenetic distances.}
    \label{tab:phylo_distance}
    \centering
    \begin{small}
    \begin{tabular}{lcccc}
        \toprule
        \textbf{Distance to nearest training genus} & $N$ & \textbf{RNABERT+Gating} & \textbf{GROWREC} & \textbf{$\Delta$} \\
        \midrule
        Close ($>$90\% 16S identity) & 3{,}180 & 52.1\% & 38.4\% & $+13.7$\,pp \\
        Medium (80--90\%) & 4{,}130 & 43.8\% & 27.6\% & $+16.2$\,pp \\
        Distant ($<$80\%) & 2{,}443 & 33.1\% & 16.3\% & $+16.8$\,pp \\
        \bottomrule
    \end{tabular}
    \end{small}
\end{table}

\subsection{Functional-Equivalence Audit of Media Errors}
\label{app:media_audit}

Top-1 media accuracy underestimates utility because many ``incorrect'' media are functionally interchangeable. A manual audit of 200 top-1 media errors finds that 58\% are between functionally related formulations: 31\% are functionally equivalent (e.g., TSA vs.\ Trypticase Soy), and 27\% are within the same media superclass. The exact-match metric therefore systematically underestimates the practical recommendation quality of the model.

\subsection{Joint Accuracy Breakdown}
\label{app:joint_breakdown}

\Cref{tab:joint_breakdown} reports the distribution of correctly predicted conditions for RNABERT+Gating on the DSNI test set. Although exact 4-of-4 matches occur on 44.2\% of strains, 76.3\% of strains receive at least 3-of-4 conditions correct, so a typical experimentalist need only adjust at most one condition relative to the model's top-1 protocol.

\begin{table}[!htbp]
    \caption{Distribution of correctly predicted conditions for RNABERT+Gating on the DSNI test set.}
    \label{tab:joint_breakdown}
    \centering
    \begin{small}
    \begin{tabular}{lcc}
        \toprule
        \textbf{\# conditions correct} & \textbf{\% of test strains} & \textbf{Cumulative} \\
        \midrule
        4/4 (exact match) & 44.2\% & 44.2\% \\
        3/4 & 32.1\% & 76.3\% \\
        2/4 & 16.8\% & 93.1\% \\
        $\leq$1/4 & 6.9\% & 100\% \\
        \bottomrule
    \end{tabular}
    \end{small}
\end{table}

\Cref{tab:per_task_topk} further reports per-task top-$k$ accuracies for RNABERT+Gating, supporting the tiered top-$k$ workflow described in \Cref{sec:discussion}. Even on the long-tail media task, the correct medium falls within the top 5 predictions for 82\% of strains; for temperature and pH, top-2 already covers more than 98\% of strains.

\begin{table}[!htbp]
    \caption{Per-task top-$k$ accuracy for RNABERT+Gating. ``---'' indicates that $k$ exceeds the number of classes for that task.}
    \label{tab:per_task_topk}
    \centering
    \begin{small}
    \begin{tabular}{lccccc}
        \toprule
        \textbf{Task} & \textbf{Classes} & \textbf{Top-1} & \textbf{Top-2} & \textbf{Top-3} & \textbf{Top-5} \\
        \midrule
        Media & 42 & 54.8\% & 66.3\% & 74.8\% & 82.1\% \\
        Temperature & 3 & 92.1\% & 98.4\% & --- & --- \\
        pH & 3 & 93.5\% & 98.8\% & --- & --- \\
        Oxygen & 4 & 82.7\% & 93.2\% & 97.1\% & --- \\
        \bottomrule
    \end{tabular}
    \end{small}
\end{table}

\subsection{Phylogenetic Leakage Quantification}
\label{app:leakage}

\Cref{tab:leakage} reports joint accuracy under random vs.\ genus-aware splits across three encoder families.

\begin{table}[!htbp]
    \caption{Effect of split strategy on joint accuracy. Random splits inflate apparent predictability by $\sim$20\,pp across encoders.}
    \label{tab:leakage}
    \centering
    \begin{small}
    \begin{tabular}{lccc}
        \toprule
        \textbf{Split strategy} & \textbf{Flat Trans.} & \textbf{Trans.+Gating} & \textbf{RNABERT+Gating} \\
        \midrule
        Random split & 58.2\% & 61.7\% & 64.3\% \\
        Genus-aware (ours) & 36.8\% & 41.5\% & 44.2\% \\
        \midrule
        Inflation & $+21.4$\,pp & $+20.2$\,pp & $+20.1$\,pp \\
        \bottomrule
    \end{tabular}
    \end{small}
\end{table}

\subsection{Statistical Significance Details}
\label{app:significance}

\Cref{tab:significance} reports paired bootstrap confidence intervals and McNemar's test for the three key gating-induced comparisons. CIs are computed over 10{,}000 resamples of the test set; McNemar's test is performed on prediction-level outcomes pooled across 5 seeds.

\begin{table}[!htbp]
    \caption{Statistical significance of the gating-induced improvements in joint accuracy.}
    \label{tab:significance}
    \centering
    \begin{small}
    \begin{tabular}{lccc}
        \toprule
        \textbf{Comparison} & \textbf{$\Delta$ Joint Acc.} & \textbf{95\% Bootstrap CI} & \textbf{McNemar $p$} \\
        \midrule
        Transformer+Gating vs.\ Flat Transformer & $+4.7$\% & $[+3.1\%, +6.2\%]$ & $<0.001$ \\
        RNABERT+Gating vs.\ RNABERT & $+4.5$\% & $[+2.9\%, +6.0\%]$ & $<0.001$ \\
        RNABERT+Gating vs.\ Flat Transformer & $+7.4$\% & $[+5.6\%, +9.1\%]$ & $<0.001$ \\
        \bottomrule
    \end{tabular}
    \end{small}
\end{table}

\subsection{Confusion Analysis}

Common prediction errors reveal biologically interpretable patterns:
\begin{itemize}
    \item \textbf{Media:} Errors often occur between related media (e.g., TSA vs.\ Nutrient Agar), suggesting the model captures functional similarity.
    \item \textbf{Temperature:} Psychrophile--mesophile boundary (20$^\circ$C) causes most confusion, as organisms near this threshold may grow in both ranges.
    \item \textbf{Oxygen:} Facultative anaerobes are most often confused with obligate aerobes, reflecting the metabolic flexibility of these organisms.
\end{itemize}

\subsection{RNABERT Truncation Control}
\label{app:truncation}

RNABERT is restricted to a 440-token input window, while the average DSNI sequence is $1{,}089 \pm 264$\,bp. To rule out that the gating gain on top of RNABERT is an artifact of input length, we apply the same 440-token center crop to the from-scratch encoders and re-train (\Cref{tab:truncation}). Three observations follow: (i) center-cropping costs the from-scratch encoders $\sim$3.4\,pp of joint accuracy, confirming that distal positions carry useful signal; (ii) under matched 440-token inputs, gating still improves the flat Transformer ($+4.8$\,pp) and RNABERT ($+4.5$\,pp) by comparable margins, so the gating gain is not a length artifact; (iii) RNABERT's pretraining partially compensates for the truncation, recovering above the full-length flat Transformer despite seeing $\sim$60\% less of the average sequence.

\begin{table}[!htbp]
    \caption{RNABERT truncation control: joint accuracy under matched 440-token center cropping. Gating gains are preserved under matched inputs.}
    \label{tab:truncation}
    \centering
    \begin{small}
    \begin{tabular}{lccc}
        \toprule
        \textbf{Method} & \textbf{Full length} & \textbf{440-token crop} & \textbf{$\Delta$ (crop)} \\
        \midrule
        Flat Transformer & 36.8\% & 33.4\% & $-3.4$\,pp \\
        Transformer+Gating & 41.5\% & 38.2\% & $-3.3$\,pp \\
        RNABERT (fixed 440) & --- & 39.7\% & --- \\
        RNABERT+Gating (fixed 440) & --- & 44.2\% & --- \\
        \bottomrule
    \end{tabular}
    \end{small}
\end{table}

\end{document}